\documentclass[twocolumn]{article}
\usepackage{cite}
\usepackage{amsmath,amssymb,amsfonts}
\usepackage{algorithmic}
\usepackage{graphicx}
\usepackage{textcomp}
\usepackage{xcolor}
\def\BibTeX{{\rm B\kern-.05em{\sc i\kern-.025em b}\kern-.08em
    T\kern-.1667em\lower.7ex\hbox{E}\kern-.125emX}}

\usepackage{caption} 

\usepackage{hyperref}

\usepackage{multirow}
\usepackage{subfig}
\usepackage{pgfplots}
\usepackage{array}
\usepackage{authblk}

\newcommand{\sref}[1]{``\ref{#1}''}

\begin{document}

\title{Non-Generative Energy Based Models\\
}

 \author[1]{Jacob Piland (jpiland@nd.edu)}

 \author[2]{Christopher Sweet (csweet1@nd.edu)}

\author[3]{Priscila Saboia (pmoreira@nd.edu)}

\author[4]{Charles Vardeman II (cvardema@nd.edu}

\author[5]{Adam Czajka (aczajka@nd.edu)}

 \affil[1,5]{Computer Science and Engineering, University of Notre Dame,
 Notre Dame, USA}
\affil[2,3,4]{Center for Research Computing, University of Notre Dame,
 Notre Dame, USA}

\maketitle

\begin{abstract}
Energy-based models (EBM) have become increasingly popular within computer vision. EBMs bring a probabilistic approach to training deep neural networks (DNN) and have been shown to enhance performance in areas such as calibration, out-of-distribution detection, and adversarial resistance. However, these advantages come at the cost of estimating input data probabilities, usually using a Langevin based method such as Stochastic Gradient Langevin Dynamics (SGLD), which 
bring additional computational costs, require parameterization, caching methods for efficiency, and can run into stability and scaling issues. EBMs use dynamical methods to draw samples from the probability density function (PDF) defined by the current state of the network and compare them to the training data using a maximum log likelihood approach to learn the correct PDF. 

We propose a non-generative training approach, Non-Generative EBM (NG-EBM), that utilizes the {\it{Approximate Mass}}, identified by Grathwohl et al. \cite{grathwohl2019classifier}, as a loss term to direct the training.
We show that our NG-EBM training strategy retains many of the benefits of EBM in 
calibration, out-of-distribution detection, and adversarial resistance, but without the computational complexity and overhead of the traditional approaches. In particular, the NG-EBM approach improves the Expected Calibration Error by a factor of 2.5 for CIFAR10 and 7.5 times for CIFAR100, when compared to traditionally trained models. 
\end{abstract}

\section{Introduction}

A neural network's prediction confidence for an event, regardless of the model's accuracy, does not typically reflect the real-world probability of the event unless trained to do so. This disconnect makes it impossible to compare predictions even from the same model. It makes it difficult for users to trust or understand the model's predictions. It also makes it difficult to combine outcomes of different models into an ensemble. These problems can be addressed by model calibration \cite{grathwohl2019classifier} that brings incompatible confidence scores to the common ground of reflecting real-world probability. Energy-based models (EBMs) \cite{lecun-06}  are generative models \cite{chapelle2006semi,dempster1977maximum} that leverage statistical approaches to improve calibration. They also have been shown \cite{grathwohl2019classifier} to improve out-of-distribution detection, where concepts such as samples from a distribution lying in the ``typical'' set have led to new metrics like {\it{Approximate Mass}}. Furthermore, adversarial resistance is improved, where simply adding EBM training can produce more robust models. However, these enhancements are bought at the cost of introducing a generative component that adds significant computational cost, complexity, stability issues and additional hyper-parameters, which is a barrier to wider utilization. 

{\bf{Contributions:}} To facilitate the scalability 
of EBMs, we investigate using an alternative method that utilizes the {\it{Approximate Mass}}, identified by \cite{grathwohl2019classifier}, as a loss term to direct the training without using generative methods to learn the correct probability density function (PDF). 
This contrasts with methods based on SGLD which use maximum log likelihood to compare training data with samples drawn using dynamical methods from the distribution defined by the current state of the network to learn the correct PDF. 
Source code for the proposed method and model weights are made available with the paper.

{\bf{Paper organization:}} First, we discuss related work and its impact on our work in Section \sref{sect:related}, 
we then briefly describe how EBMs function in Section \sref{sect:ebms} with a focus on the generative step. Then we explain the basis and implementation of our NG-EBM method in Section \sref{sect:nongen}. We introduce the results of a state of the art generative EBM to which we directly compare the NG-EBM experimentally in section \sref{sect:conclusions}. Limitations and further work ideas are provided in sections \sref{sect:limitations} and \sref{sect:conclusions}, respectively.

\section{Related Work}
\label{sect:related}
Our work is based on two previously published methods JEM \cite{grathwohl2019classifier} and Jem++ \cite{JEMpp}. 
In \cite{grathwohl2019classifier}, the Authors use the EBM approach to develop a new method, the Joint Energy Model (JEM) which uses Stochastic Gradient Langevin Dynamics (SGLD) \cite{Welling_bayesianlearning} to generate representative samples to estimate the energy and maximum likelihood estimation (MLE) for training. SGLD, and Langevin methods in general, are popular sampling technique which have several drawbacks for machine Learning due to computational cost, sensitive parametrization and stability issues. During our experiments only around 50\% of JEM training runs completed successfully with remaining failed runs exhibiting unbounded growth in output from the SGLD simulation of Eq.(\ref{eqn:sgld}), a know stability issue with dynamical approaches where work is in progress \cite{https://doi.org/10.48550/arxiv.2012.01316}. 

In addition to stability issues, compromises are made for efficiency, for example by re-using a large number of cached SGLD data points for initial conditions since JEM is restricted to 20 integration steps per training batch for efficiency. In JEM++ \cite{JEMpp} the authors optimize the JEM method and note that one of the method's weaknesses is starting the SGLD from a random input that requires eitherlong generative simulation or complex caching schemes. Another weaknesses is the stability problem noted above and the Authors use a two player min-max problem for their solution.

Our contribution is to further refine the method, based on the observations and results from both JEM and JEM++, by removing the generative requirement and hence the stability and initialization problems.

\section{Energy Based Models}
\label{sect:ebms}
Energy based models \cite{lecun-06} are based on the idea that the probability density $p(x)$ for data points $x$ in the domain ($x \in \mathbb{R}^\mathrm{D}$) can be expressed as,
\begin{align}
\label{eqn:probdense}
p_{\theta}(x) = \frac{e^{-E_{\theta} (x)}}{Z_{\theta}}, \qquad Z_{\theta}=\int e^{-E_{\theta} (x)}dx.
\end{align}
Where $E_{\theta}(x)$, defined as the ``energy'', is a scalar function parameterized by $\theta$, the deep neural network (DNN) characteristics and weights, and $Z_{\theta}$. Defining $E_{\theta} (x)$ for a DNN, $p_{\theta}(x)$ allows us to learn an underlying data distribution by analyzing the observed data. EBMs have become increasingly popular within computer vision in recent years, commonly being applied for various generative image modeling tasks \cite{Du_ebm_2019,8579052,9157612,Nijkamp2019LearningNN,Nijkamp2019OnTA}.

For most choices of $E_{\theta} (x)$, one cannot compute or even reliably estimate $Z_{\theta}$, which means estimating normalized densities is intractable and standard maximum likelihood estimation (MLE) of the parameters, $\theta$, is not straightforward. For techniques that seek to maximize the log-likelihood we usually resort to methods like MCMC \cite{Younes1989,Hinton_2002} to draw samples from $p_{\theta}(x)$. MCMC requires numerical methods that take multiple steps to sample from the PDF, which adds computational cost and complexity to the method in addition to stability issues as the numerical method must balance the competing requirements of large learn rate for efficiency and small learning rate for accuracy. 

Machine learning classification problems with $K$ classes is typically addressed using a parametric function, $f_\theta : \mathbb{R}^D \mapsto \mathbb{R}^K$, which maps each data point $x \in \mathbb{R}^D$ to $K$ real-valued numbers known as logits. These logits are used to parameterize a categorical distribution using the so-called Softmax transfer function,
\begin{align}
\label{eqn:softmax}
p_\theta(y | {x}) = \frac{e^{f_\theta(x)_y}}{\sum_{\hat{y}}e^{f_\theta(x)_{\hat{y}}}},
\end{align}
where $f_\theta(x)_y$ indicates the $y^{\mathrm{th}}$ index of $f_\theta(x)$, i.e. the logit corresponding the the $y^{\mathrm{th}}$ class label.

Now we connect this categorial distribution to the concept of energy.
From the definition of conditional probability we have,
\begin{align}
\label{eqn:condprob}
p_\theta(y | x) = \frac{p_\theta(x, y)}{p_\theta(x)}.
\end{align}
Hence, for a given $x$ and substituting Eq. (\ref{eqn:condprob}) into Eq. (\ref{eqn:softmax}) we can show,
\begin{align}
\label{eqn:frelatedtopx}
p_\theta(x, y) \propto e^{f_\theta(x)_y}, \qquad p_\theta(x) \propto \sum_{\hat{y}}e^{f_\theta(x)_{\hat{y}}}.
\end{align}
Then note that by substituting the energy function,
\begin{align}
E_\theta(x) = -\log\left(\sum_{y} e^{f_\theta(x)_{y}}\right),
\label{eqn:energy}
\end{align}
into the $p_\theta(x)$ equation in Eq. (\ref{eqn:frelatedtopx}) yields Eq. (\ref{eqn:probdense}) as required.

These parameters must be estimated and maximum likelihood estimation (MLE) is a popular method, given some observed data. For EBM we note that the derivative of the log-likelihood for a single example $x$ with respect to $\theta$ can be expressed as,
\begin{align}
\label{eqn:loglikelyhood}
\frac{\partial \log p_{\theta}(x)}{\partial \theta} = \mathbb{E}_{p_\theta(\tilde{x})}\left[\frac{\partial E_\theta(\tilde{x})}{\partial \theta}\right]-\frac{\partial E_\theta(x)}{\partial \theta},
\end{align}
where the expectation is over the model distribution. However, it is not easy to draw samples from $p_\theta(x)$, so we use methods such as MCMC for this gradient estimator. Some of the earliest EBMs have used this approach, an example being Restricted Boltzmann Machines \cite{Hinton06}, and more recent work uses this method to train large-scale EBMs \cite{nijkamp2019anatomy,nijkamp2019learning,xie2016theory,Du_ebm_2019}. 

Other recent work has approximated the expectation in Eq. (\ref{eqn:loglikelyhood}) using a sampler based on Stochastic Gradient Langevin Dynamics (SGLD) \cite{Welling_bayesianlearning} which draws samples following,
\begin{align}
\label{eqn:sgld}
\nonumber
x'_0\sim p_0(x), \qquad x'_{i+1}=x'_i-\frac{\alpha}{2}\frac{\partial E_\theta(x'_i)}{\partial x'_i}+\epsilon, \\
\qquad \epsilon \sim \mathcal{N}(0,\alpha),
\end{align}
where $p_0(x)$ is typically a Uniform distribution over the input domain and the step-size $\alpha$ should be decayed following a polynomial schedule. The expectation over the model distribution is then approximated by the average of the samples from Eq. (\ref{eqn:sgld}), usually a number equal to the batch size, yielding loss function over the batch, 
\begin{align}
\label{eqn:loss}
L^{\mathrm{EBM}}_\theta(X,X') = \sum_{x'\in X'}E_\theta(x')-\sum_{x\in X}E_\theta(x),
\end{align}
where $X'$ is the set of generative samples $x'$, the results from propagating Eq. (\ref{eqn:sgld}) and $X$ is the batch from the training set. The partial derivative of $L^{\mathrm{EBM}}_\theta$ w.r.t. $\theta$ then approximates Eq. (\ref{eqn:loglikelyhood}) during back-propagation.

In the above Section \sref{sect:related} we discussed JEM \cite{grathwohl2019classifier} as an example of an EBM.

\section{Non-generative EBM}
\label{sect:nongen}
JEM uses SGLD to draw samples with which one can estimate the expected value for the energy. From random initial conditions we might expect these samples to converge to their statistically correct values given sufficient steps. Since the cost of calculating $\partial E_{\theta}/\partial x$ requires back propagation this method is computationally expensive. As discussed above, Grathwohl et al. \cite{grathwohl2019classifier} mitigates this cost by taking few steps, around 20 for their published results, but using a caching technique where the results of the SGLD are stored and, at each step, 95\% of samples are pulled from the cache and  5\% are generated randomly. This allows the cached samples to evolve over multiple SGLD runs.

In JEM++, Yang et al. \cite{JEMpp} make a number of observations before proposing a different approach to generating the samples. Firstly, similar to Nijkamp et al. \cite{nijkamp2019anatomy} who have found the insignificance of noise term in the SGLD sampling Eq.(\ref{eqn:sgld}), the empirical study by Yang et al. \cite{JEMpp} also confirms this observation. Secondly, they state that, assuming  the  convergence  can  be achieved, the objective of the SGLD sampling Eq.(\ref{eqn:sgld}) is to solve the optimization problem  $x^{*}= \underset{x}{\mathrm{argmax}} ~E_{\theta}(x)$ approximately. Using this information they proceed to train by using a two player min-max problem with considerable success.

Following the logic of Yang et al. \cite{JEMpp}, if we take the SGLD Eq.(\ref{eqn:sgld}) and remove the noise term as they propose, at step $i$ we have,
\begin{align}
\label{eqn:sgldnonoise}
 x'_{i+1}=x'_i-\frac{\alpha}{2}\frac{\partial E_\theta(x'_i)}{\partial x'_i},
\end{align}
which takes the form of a minimizer, assuming convergence to an approximate minimum we expect the energy derivative magnitude,
\begin{align}
\label{eqn:sgldminima}
EGM_{\theta}( x'_i)=\left\lVert\frac{\partial E_\theta(x'_i)}{\partial x'_i}\right\rVert_2,
\end{align}
to be small with  $EGM_{\theta}( x'_i)< \eta$ for some small $\eta$ and large $i$.

We should note that, Grathwohl et al. \cite{grathwohl2019classifier} proposed a metric called {\it{Approximate Mass}} for ``out-of-distribution'' OOD detection,
\begin{align}
\label{eqn:approxmass1}
s_\theta(x) = -\left\lVert\frac{\partial \log p_\theta(x)}{\partial x}\right\rVert_2.
\end{align}
After substituting Eq.(\ref{eqn:probdense}) into Eq.(\ref{eqn:approxmass1}) we get a similar equation (up to sign) to Eq.(\ref{eqn:sgldminima}),
\begin{align}
\label{eqn:approxmass2}
s_\theta(x)  = -\left\lVert\frac{\partial E_\theta(x)}{\partial x}\right\rVert_2.
\end{align}

From Grathwohl et al. \cite{grathwohl2019classifier} the explanation for Approximate Mass is this as follows: ``Real samples from a distribution lie in what is known as the ``typical'' set. This is the area of high probability mass. A single point may have high density but if the surrounding areas have very low density, then that point is likely not in the typical set and therefore likely not a sample from the data distribution. For a high-likelihood data point outside of the typical set, we expect the density to change rapidly around it, thus the norm of the gradient of the log-density will be large compared to examples in the typical set (otherwise it would be in an area of high mass).''

\begin{figure*}[ht]
\centering
\input{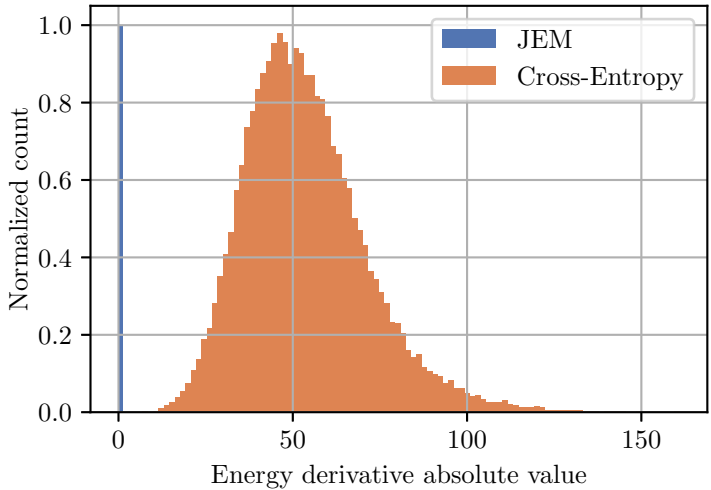}
\caption{Histogram displaying the normalized count for the energy derivatives for the samples of the CIFAR10 training set using Cross-Entropy and JEM.}
\label{fig:energy_derivative}
\end{figure*}

From this, we could reasonably expect that both the generated samples and the training set to belong to the ``typical set'' and the gradient of the log-density/energy should be small for both $x'$ and $x$. From Eq.(\ref{eqn:sgldminima}) this is true for the generated samples $x'$ and so should hold for the training set $x$. Experimental data confirms that this is the case in Figure \ref{fig:energy_derivative} where we compare JEM to a model trained only with Cross-Entropy.

Based on the observations from prior work \cite{grathwohl2019classifier,JEMpp}, we propose a simplified method, NG-EBM, that removes the SGLD sampling/MLE terms in Eq.(\ref{eqn:loss}) and adds a term to the loss function that favours low magnitude Energy derivatives for the training data. This method utilizes Eq.(\ref{eqn:sgldminima}) multiplied by $-1$ (similar to {\it{Approximate Mass}}) as a loss function term with the training data $x$ substituted for the generated samples $x'$ as discussed above,
\begin{align}
\label{eqn:lossngebm3}
L^{\mathrm{NG-EBM}}_\theta(X) = -\beta\sum_{x\in X}\left\lVert\frac{\partial E_\theta(x)}{\partial x}\right\rVert_2,
\end{align}
for scaling coefficient $\beta$. Note that the complete loss function will include a Cross-Entropy term which will have an additional scaling factor $\gamma$ such that $\beta + \gamma = 1$. For our experiments we choose $\beta = 0.5$ and $\gamma=0.5$.

We show that, while we do not claim that NG-EBM captures all of the more subtle characteristics of EBMs, it does perform equally as well as JEM for calibration, adversarial resistance and out of distribution detection without the computational cost and stability issues inherent to SGLD.

\section{Experiments and Results}
\label{sect:experiments}
Similar to the experiments proposed by \cite{grathwohl2019classifier}, we explore the performance of our formulation of NG-EBM, with comparison to their Joint Energy Models (JEM), in the areas of calibration, out-of-distribution detection (OOD) and robustness against adversarial attacks. OOD is particularly important since the Approximate Mass is closely related to the underlying behaviour of both EBMs and NG-EBM.

The purpose of our NG-EBM is {\bf{not to outperform JEM}} in terms of accuracy, adversarial resistance, calibration and out-of-distribution detection, but to match the state of the art while reducing complexity and computational cost as seen in Table \ref{table:cost}. We see a 6-fold reduction in cost for NG-EBM compared to JEM. For these experiments we trained five models for each of Cross-Entropy, JEM and NG-EBM. 

\begin{table*}[ht]
\centering
\begin{tabular}{ l l l l } 
\hline 
Model & Number & Training time per model (hours) & Total cost (Kg C$0_2$)\\
\hline 
Cross-Entropy  & 5 & 7.5 & 17.1 \\ 
JEM  & 5 & 85 & 192.8 \\ 
NG-EBM  & 5 & 15 & 34.1 \\ 
\hline \\
\end{tabular}
\caption{C$0_2$ cost for experiments using RTX 3090 GPU. Calculated at \url{https://mlco2.github.io/impact/} with default 0.432Kg/kWh.}
\label{table:cost}
\end{table*}

Following \cite{grathwohl2019classifier} and using code from that repository \url{https://github.com/wgrathwohl/JEM} under an Apache-2.0 license (that we have modified), we train all models with the Adam optimizer \cite{kingma2014adam} for 150 epochs through the dataset using a staircase decay schedule for the learning rate as reported in \cite{grathwohl2019classifier}. All architectures used are based on Wide Residual Networks, Wide-ResNet \cite{Zagoruyko_wide_2016}, where we have removed batch-normalization to ensure that our models’ outputs are deterministic functions of the input.

We utilize the same datasets as \cite{grathwohl2019classifier}. For training the models we use both CIFAR10 and CIFAR100 datasets \cite{Krizhevsky_2009_17719} and for testing out-of-distribution detection we additionally use SVHN \cite{SVHN_37648}. The JEM authors also use an interpolated version of CIFAR10, CIFAR10 Int., where the the mean of the present and previous batch is taken and a random perturbation, sampled from $\mathcal{N}(0,0.001)$, is added. All these datasets were obtained via Torchvision version 0.12.0 \cite{10.1145/1873951.1874254} and metrics are calculated using the designated ``test'' datasets. 


\subsection{Calibration Experiment}
A classifier is considered calibrated if its predictive confidence, $\max_y p_\theta(y|x)$, aligns with its classification rate. This can be important feature for a model when deployed in real-world scenarios. When a calibrated classifier predicts a given label $y$ with a confidence value, say 0.7, it should have a corresponding chance of being correct, in this case 70\%. Whereas recent models have improved in accuracy this trend has not transferred to calibration, which has actually worsened \cite{Guo_calibration_2017}. In \cite{grathwohl2019classifier} the authors observed that JEM dramatically improves calibration while retaining high accuracy. 

\begin{figure*}[ht]
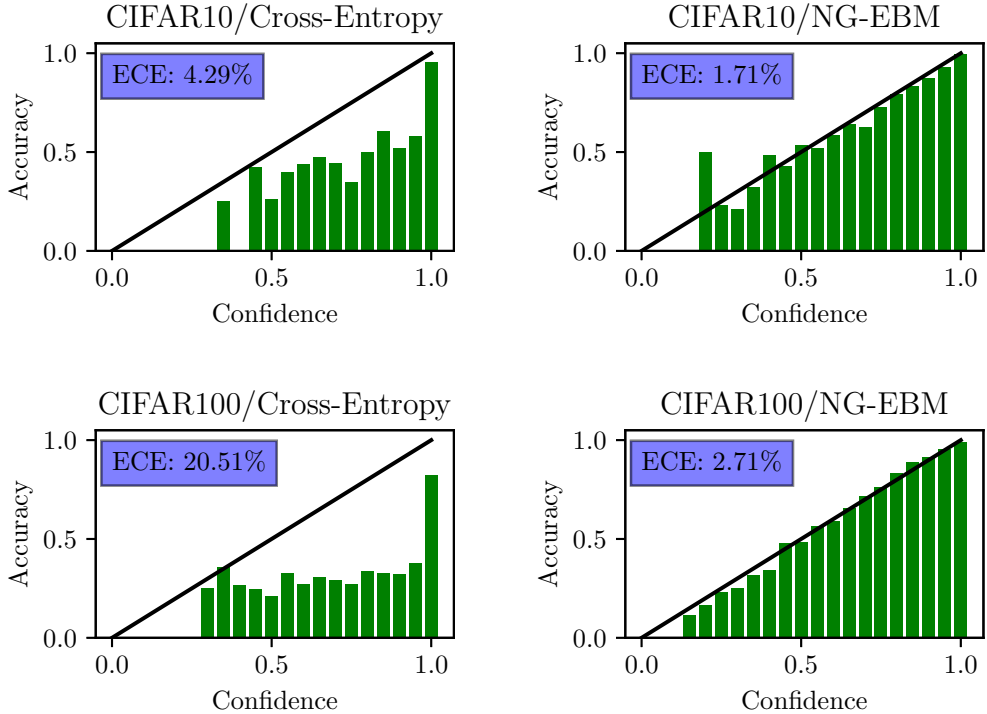

\centering
    \subfloat{{ \input{plots/calibration_XE_cifar10_ce.pgf} }}
    \subfloat{{ \input{plots/calibration_NG-EBM_cifar10.pgf} }} \\
    \subfloat{{ \input{plots/calibration_XE_cifar100_ce.pgf} }}
    \subfloat{{ \input{plots/calibration_NG-EBM_cifar100.pgf} }}
\caption{Calibration comparison between methods for CIFAR10 models (top) and CIFAR100 models (bottom) trained using both Cross-Entropy and NG-EBM as their loss functions. ECE is the Expected Calibration Error \cite{Guo_calibration_2017}.  Note that these are specific examples chosen at random for illustration from the 15 trained models, statistics over the full set are in Table \ref{table:cal}.}
\label{fig:calibration}
\end{figure*}

In Figure \ref{fig:calibration} we observe that our NG-EBM retains this feature for both the CIFAR10 and CIFAR100 data sets and is a considerable improvement on Cross-Entropy in both cases, improving the Expected Calibration Error (ECE) \cite{Guo_calibration_2017} by a factor of 2.5 for CIFAR10, 7.5 times for CIFAR100 and being closer to the ideal of monotonically increasing for both these examples. ECE results for datasets CIFAR10 and CIFAR100, with models trained using Cross-Entropy, JEM, and NG-EBM can be seen in Table \ref{table:cal}. 

\begin{table*}[ht]
\centering
\begin{tabular}{ l l l l } 
\hline
& Cross-Entropy & JEM & NG-EBM\\
\hline
CIFAR10  & 5.0$\pm0.1$\% & 4.1$\pm1.1$\% & 1.8$\pm0.8$\% \\ 
CIFAR100  & 22.7$\pm0.4$\% & 8.1$\pm3.2$\% & 3.3$\pm1.7$\% \\ 
\hline \\
\end{tabular}
\caption{Expected Calibration Error Results for datasets CIFAR10 and CIFAR100 each with models trained using Cross-Entropy, JEM and NG-EBM. Standard errors are calculated over five models.}
\label{table:cal}
\end{table*}


We note that, compared to other calibration methods such as Platt scaling \cite{Guo_calibration_2017}, both NG-EBM and JEM requires no additional training data.

\subsection{Out-Of-Distribution Detection Experiment}
\label{sect:ood}
For this experiment models trained on CIFAR10 were scored against SVHN, CIFAR10 Int. and CIFAR100 and models trained on CIFAR100 were scored against SVHN, CIFAR10 Int. and CIFAR10.

We can consider out-of-distribution (OOD) detection as a binary classification problem, where the model is required to produce a score,
\begin{align}
\label{eqn:ood}
s_\theta(x) \in \mathbb{R},
\end{align}
where $x$ is the query, and $\theta$ is the set of learnable parameters. Our expectation is that the scores for in-distribution examples are higher than that out-of-distribution examples. Following best practices, we use the area under the receiver-operating curve (AUROC) \cite{Hendrycks_ood_2016} as our threshold-free metric. We compare three OOD detection approaches when comparing the NG-EBM method to JEM \cite{grathwohl2019classifier} and models trained with to Cross-Entropy loss.

\subsubsection{Input Density}
For methods that can estimate the input density $p_\theta(x)$ it is common to consider examples with low likelihood to be OOD.
In practice EBMs cannot estimate the (calibrated) input density due to the intractable normalizing constant but can compute relative input density \cite{Nijkamp2019OnTA,pmlr-v9-gutmann10a,Tieleman_2008}. However, \cite{Nalisnick_generative_2018} showed that tractable deep generative models such as \cite{Kingma_generative_2018} and \cite{Salimans_2017} can assign higher densities to OOD examples than in-distribution examples. More recent work from \cite{Nalisnick_ood_2019} shows examples where the densities of an OOD dataset are completely indistinguishable from the in-distribution set but, conversely \cite{Du_ebm_2019} have shown that the likelihoods from EBMs can be reliable in use for predicting for OOD inputs. Quantitative results for our NG-EBM method, JEM and models trained with to Cross-Entropy loss can be found in Table \ref{table:ood}. JEM and our NG-EBM give similar results but close to the minimum "acceptable" value of 0.7 for AUROC, supporting the observations of \cite{Nalisnick_ood_2019}. As can be seen in Table \ref{table:score_dist} column 2, both JEM and NG-EBM consistently assigns higher likelihoods to in-distribution data than OOD data.

\subsubsection{Predictive Distribution}
Successful approaches to OOD prediction have utilized a classifier’s predictive distribution \cite{Gal_uncertain_2015,Wang_bayesian_2018,Liang_ood_2017}. In this case the score is the maximum prediction probability: $s_\theta(x) = \max_y p_\theta(y|x)$ \cite{Hendrycks_ood_2016}. It has also been demonstrated that OOD performance using this score is highly correlated with a model’s classification accuracy. Results can be seen in Table \ref{table:ood} (middle) with similar results for Wide-ResNet with Cross-Entropy, JEM and NG-EBM.

\subsubsection{Approximate Mass}
\label{sect:approxmass}
It has been observed that likelihood may not be enough for OOD detection in high dimensions \cite{Nalisnick_ood_2019}. The rational of Grathwohl et al. \cite{grathwohl2019classifier} is detailed in Section \sref{sect:nongen} but considers that real samples from a distribution lie in what they call the ``typical'' set which has an area of high probability ``mass''. Based on their 
intuition that a high-likelihood datapoint outside of the typical set would see the density change rapidly in its neighborhood, 
the Authors proposed a new metric which they termed Approximate Mass and represents our additional loss term from Section \sref{sect:nongen},
\begin{align}
\label{eqn:approxmass}
s_\theta(x) = -\left\lVert\frac{\partial E_\theta(x)}{\partial x}\right\rVert_2.
\end{align}
Since this addition to the loss Function should minimize the term, we expect this to achieve good results as seen in Table \ref{table:ood} (bottom) which compares the AUROC values for OOD. The NG-EBM and JEM results are very similar. For both JEM and NG-EBM, we find this predictor outperforms their respective model’s likelihoods as seen in Table \ref{table:score_dist} column 3.

\begin{table*}[ht]
\centering
\begin{tabular}{ l l l l l } 
\hline
$s_\theta(x)$ & Model & SVHN & CIFAR10 Int. & CIFAR100\\
\hline\rule{0pt}{2ex} 
$\log p(x)$  & \multirow{2}{4em}{JEM NG-EBM} & 0.67$\pm0.03$ & 0.65$\pm0.03$ & 0.67$\pm0.01$\\ 
&  & 0.73$\pm0.04$ & 0.63$\pm0.03$ & 0.69$\pm0.03$\\ 
\hline\rule{0pt}{2ex} 
$\max_y p_\theta(y|x)$  & \multirow{3}{6em}{Wide-ResNet\\ JEM \\ NG-EBM} & 0.93$\pm0.01$ & 0.75$\pm0.01$ & 0.85$\pm0.01$\\ 
&  & 0.88$\pm0.02$  & 0.75$\pm0.02$ & 0.84$\pm0.01$\\ 
&  & 0.88$\pm0.02$  & 0.78$\pm0.03$ & 0.84$\pm0.02$\\ 
\hline\rule{0pt}{3ex}
$-\left|\frac{\partial E_\theta(x)}{\partial x}\right|$  & \multirow{2}{4em}{JEM NG-EBM} & 0.70$\pm0.07$ & 0.83$\pm0.01$ & 0.74$\pm0.04$\\[-1ex] 
&  & 0.71$\pm0.02$ & 0.81$\pm0.02$ & 0.76$\pm0.01$\\ 
\hline \\
\end{tabular}
\caption{AUROC values obtained for OOD detection problem. Models trained on CIFAR10. Standard errors are calculated over 5 models.}
\label{table:ood}
\end{table*}

\begin{table*}[ht]
\centering
\begin{tabular}{>{\centering\arraybackslash} m{1.5cm} 
                >{\centering\arraybackslash} m{5cm}
                >{\centering\arraybackslash} m{5cm}}
\hline\rule{0pt}{3ex}
Method & $\log p_\theta(x)$ & $-\left|\frac{\partial E_\theta(x)}{\partial x}\right|$ \\[1ex]
\hline \\
JEM & \input{plots/score_sgld_px.pgf} & \input{plots/score_sgld_pxgrad.pgf} \\
\hline \\
NG-EBM & \input{plots/score_ngebm_px.pgf} & \input{plots/score_ngebm_pxgrad.pgf} \\
\hline \\
\end{tabular}
\caption{Histograms for OOD (Out Of Distribution) detection. All models trained on CIFAR10. Blue corresponds to the score on (in-distribution) CIFAR10, and red corresponds to the score on the OOD dataset. Note: histogram heading is the AUROC score from Table \ref{table:ood} and axis labels have been removed for clarity (method value in the column heading is the x-axis), OOD is dependent on histogram overlap. For each graph the x-axis is the metric listed above the column and the y-axis is the density of probability both scaled to fit.}
\label{table:score_dist}
\end{table*}

\subsection{Robustness Against Adversarial Attacks}
Adversarial robustness is commonly measured by generating perturbation-based adversarial examples using an $L_p$-norm, for some $p$, as a constraint \cite{Goodfellow_2014}. Adversarial inputs $\hat{x}$ are generated by adding perturbation $\eta$ to a known input $x$ such that $\hat{x} = x + \eta$ and which change a model’s prediction subject to $||\hat{x}-x||_p < \epsilon$. The examples $\hat{x}$ exploit perturbations to which the model is overly sensitive but with no implication that they reside within areas of high density according to the model distribution.

\begin{figure*}[!ht]
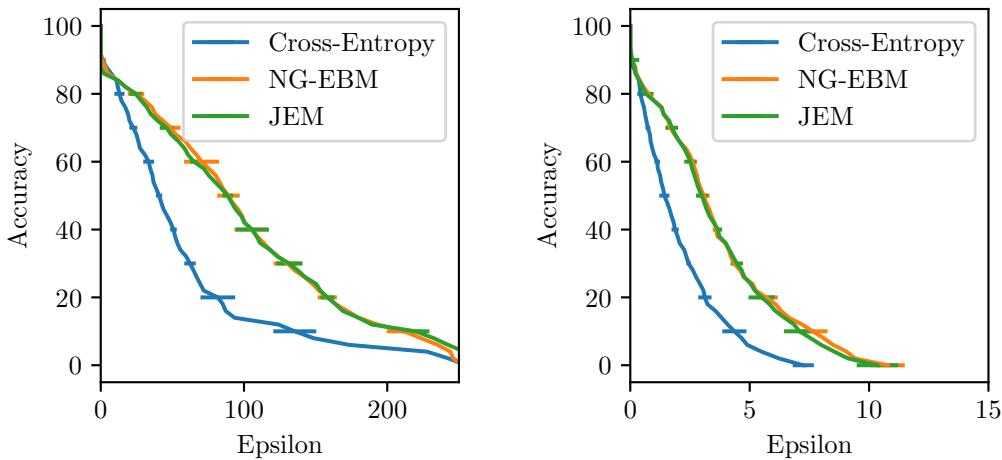

\centering
    \subfloat{{ \input{plots/adversarials_cifar10_l2_ce.pgf} }}
    \subfloat{{ \input{plots/adversarials_cifar10_linf_ce.pgf} }}
\caption{Comparison between PGD-driven adversarial attacks for CIFAR10 models trained with NG-EBM, JEM and Cross-Entropy, using the $L_2$ norm (left plot) and $L_\infty$ norm (right plot) for the attack. The higher accuracy for a given level of perturbation (epsilon), the better (more robust) the model. NG-EBM and JEM provide similar resistance to adversarial attacks, which is superior to the Cross-Entropy-trained model. Horizontal bars represent the standard error over five models.}
\label{fig:adv_cifar10}
\end{figure*}

To measure perturbation robustness, we run white-box Projected Gradient Descent (PGD) adversarial attacks on our CIFAR10-trained models using Foolbox version 1.8 \cite{Rauber_2017} with both $L_2$ and $L_\infty$ norms. As seen in Fig. \ref{fig:adv_cifar10}, both the proposed NG-EBM and JEM \cite{grathwohl2019classifier} show enhanced robustness against PGD-driven adversarial attacks, when compared to the model trained with Cross-Entropy (with JEM and NG-EBM having similar results
for various amounts of perturbation (epsilon)).




\section{Limitations}
\label{sect:limitations}
In comparing the proposed NG-EBM and JEM methods we have largely followed the methodology of \cite{grathwohl2019classifier} and thus have only fully looked at two training datasets (CIFAR10 and CIFAR100) and a single CNN architecture (Wide-ResNet). We perform some limited training with ILSVRC2012 to test scalability. To confirm the generalization capabilities of the proposed method, other large datasets and other model architectures may be incorporated.

Experimentally, introducing the {\it{Approximate Mass}} into the loss function provides similar behaviours for calibration, adversarial resistance and out-of-distribution detection as the JEM EBM. The NG-EBM method may not reproduce other important characteristics of EBMs that we have not tested and further research into this potential limitation is required.

Computationally, are not aware of any limitations of our method when compared to EBMs using MCMC and SGLD techniques. However, in common with generative methods such as SGLD our loss function requires a second derivative of the energy term to be calculated that effectively doubles the storage requirements for the data batch, this is particularly relevant for GPU use. 

\section{Conclusions and Further Work}
\label{sect:conclusions}
In this work we have presented NG-EBM, a novel approach to incorporating some of the the key dynamical features from generative methods into a simple addition to the Loss Function that removes the need to run expensive and potentially unstable MCMC/SGLD computations. Our work is enabled by recent work in scaling techniques for training EBMs to high dimensional data and in particular the work of \cite{grathwohl2019classifier}. 

Our experiments followed the methodology proposed by \cite{grathwohl2019classifier} with the intention of comparing the results of our method with a well structured EBM, where the Authors had provided access to their code so that we could first reproduce their results. Establishing the behaviour of NG-EBM with more recent NN architectures and more complex datasets would be a natural extension to this work.

Source code is available at \url{https://github.com/nd-crane/joint-energy-models}.

\pagebreak

\bibliographystyle{plain}
\bibliography{ng-ebm_bibliography.bib}

\end{document}